\documentclass{article}

\usepackage[preprint, nonatbib]{neurips_2025} 
\usepackage[numbers]{natbib}

\usepackage[utf8]{inputenc} 
\usepackage[T1]{fontenc}    
\usepackage[hidelinks]{hyperref}       
\usepackage{url}            
\usepackage{booktabs}       
\usepackage{amsfonts}       
\usepackage{amsmath}        
\usepackage{nicefrac}       
\usepackage{microtype}      
\usepackage{xcolor}         
\usepackage{pgfplots}       
\usepackage{pgfplotstable}
\pgfplotsset{compat=1.18} 
\usepgfplotslibrary{groupplots}
\usepackage[utf8]{inputenc}
\usepackage{framed}
\usepackage{caption}
\usepackage{listings}        
\usepackage{enumitem}
\setlist{topsep=2pt, itemsep=1pt, parsep=0pt}
\makeatletter
\renewcommand{\paragraph}{\@startsection{paragraph}{4}{\z@}%
  {1.2ex \@plus 0.5ex \@minus .2ex}{-1em}{\normalsize\bf}}
\renewcommand{\section}{\@startsection{section}{1}{\z@}%
  {-1.3ex \@plus -0.4ex \@minus -0.2ex}{0.7ex \@plus 0.2ex}{\large\bf\raggedright}}
\renewcommand{\subsection}{\@startsection{subsection}{2}{\z@}%
  {-1.1ex \@plus -0.4ex \@minus -0.2ex}{0.4ex \@plus 0.1ex}{\normalsize\bf\raggedright}}
\renewcommand{\subsubsection}{\@startsection{subsubsection}{3}{\z@}%
  {-0.9ex \@plus -0.3ex \@minus -0.2ex}{0.3ex \@plus 0.1ex}{\normalsize\bf\raggedright}}
\makeatother

\title{Learning on the Job: Continual Learning from Deployment Feedback for Frozen-Weights Agents}

\author{%
  Valentin~Tablan\thanks{Main contact} \\
  The Memory Company\\
  \texttt{valentin@memco.ai} \\
  \And
  Scott~Taylor \\
  The Memory Company \\
  \And
  Kristoffer~Bernhem \\
  The Memory Company \\
}

\begin{document}

\maketitle

\begin{abstract}
AI agents encounter learning opportunities in every episode they run, and discard nearly all of them: the underlying models are frozen at deployment, so an agent that resolves a difficult request today starts from zero when it recurs tomorrow. Yet ordinary operation already produces feedback, in the form of outcome verdicts and after-the-fact corrections. We show that this feedback is a sufficient signal for continual learning when the frozen model is paired with an external memory that distils each episode into retrievable natural-language rules. On the banking domain of $\tau$-bench, against a static-RAG control retrieving over the complete policy corpus, learning from the one-bit outcome verdict lifts single-trial success to $1.6\times$ the baseline, and learning from corrections to $2.6\times$, converting 22 of the 84 tasks the baseline never solves. The result spans the deployment spectrum, measured on Mistral Large, an open-weights model that organisations with data sovereignty requirements can self-host, and replicated on a frontier model, Claude Sonnet 5. The accumulated memory also transfers: each model, reading the store built by the other, rises above its own no-memory baseline. The harness, protocol, and data are released.
\end{abstract}

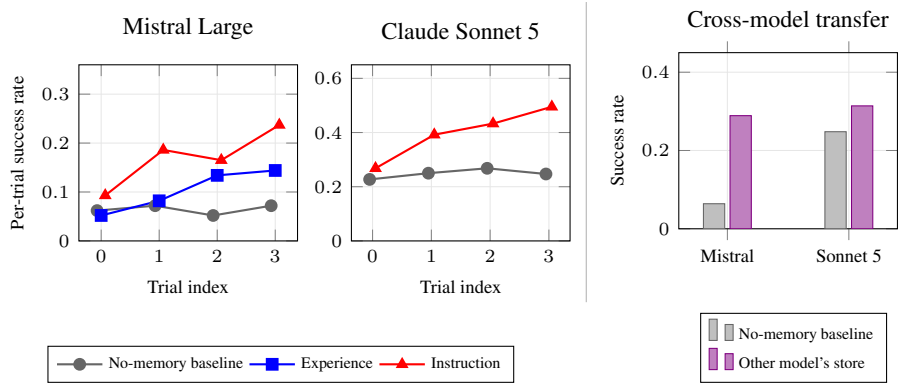
\begin{figure}[h!]
  \centering
  \begin{tikzpicture}
    \begin{axis}[
        width=0.32\textwidth, height=0.28\textwidth,
        title={Mistral Large}, title style={font=\small},
        xlabel={Trial index}, ylabel={Per-trial success rate},
        label style={font=\scriptsize}, tick label style={font=\scriptsize},
        xtick={0,1,2,3}, ymin=0, ymax=0.36,
        legend to name=leg:curvepanels, legend columns=-1,
        legend cell align=left, legend style={font=\tiny},
        grid=major, grid style={line width=0.2pt, draw=gray!20},
      ]
      \addplot[mark=*, color=black!60, thick] coordinates {
        (-0.07,0.062) (0.93,0.072) (1.93,0.052) (2.93,0.072)};
      \addlegendentry{No-memory baseline}
      \addplot[mark=square*, color=blue, thick] coordinates {
        (0,0.052) (1,0.082) (2,0.134) (3,0.144)};
      \addlegendentry{Experience}
      \addplot[mark=triangle*, color=red, thick] coordinates {
        (0.07,0.093) (1.07,0.186) (2.07,0.165) (3.07,0.237)};
      \addlegendentry{Instruction}
    \end{axis}
  \end{tikzpicture}%
  \hspace{1mm}%
  \begin{tikzpicture}
    \begin{axis}[
        width=0.32\textwidth, height=0.28\textwidth,
        title={Claude Sonnet 5}, title style={font=\small},
        xlabel={Trial index},
        label style={font=\scriptsize}, tick label style={font=\scriptsize},
        xtick={0,1,2,3}, ymin=0, ymax=0.65,
        legend pos=north west, legend cell align=left,
        legend style={font=\tiny},
        grid=major, grid style={line width=0.2pt, draw=gray!20},
      ]
      \addplot[mark=*, color=black!60, thick] coordinates {
        (-0.05,0.227) (0.95,0.250) (1.95,0.268) (2.95,0.247)};
      \addplot[mark=triangle*, color=red, thick] coordinates {
        (0.05,0.268) (1.05,0.392) (2.05,0.433) (3.05,0.495)};
    \end{axis}
  \end{tikzpicture}%
  \hspace{2mm}{\color{gray!60}\vrule width 0.4pt}\hspace{2mm}%
  \begin{tikzpicture}
    \begin{axis}[
        width=0.32\textwidth, height=0.28\textwidth,
        title={Cross-model transfer}, title style={font=\small},
        ybar, bar width=8pt,
        xlabel={\phantom{Trial index}},
        ylabel={Success rate},
        label style={font=\scriptsize}, tick label style={font=\scriptsize},
        symbolic x coords={Mistral, Sonnet 5},
        xtick=data, ymin=0, ymax=0.45,
        enlarge x limits=0.4,
        legend to name=leg:transferpanel, legend columns=1,
        legend cell align=left, legend style={font=\tiny},
        grid=major, grid style={line width=0.2pt, draw=gray!20},
      ]
      \addplot[ybar, draw=black!60, fill=black!25] coordinates {
        (Mistral,0.064) (Sonnet 5,0.248)};
      \addlegendentry{No-memory baseline}
      \addplot[ybar, draw=violet, fill=violet!50] coordinates {
        (Mistral,0.289) (Sonnet 5,0.314)};
      \addlegendentry{Other model's store}
    \end{axis}
  \end{tikzpicture}

  \smallskip
  \makebox[0.65\textwidth][c]{\ref{leg:curvepanels}}\makebox[0.32\textwidth][c]{\ref{leg:transferpanel}}
  \caption{Left two panels: per-trial success rate under continual learning for Mistral Large and Claude Sonnet 5 (note the different y scales). \textbf{No-memory baseline} is static RAG, \textbf{Experience} is learning from success/failure feedback, \textbf{Instruction} is learning from corrections. The no-memory and learning conditions start at parity on the cold trial, the baselines stay flat, and the learning conditions rise as the memory store fills. Right: cross-model transfer; each consumer's success rate reading the other model's frozen store, against its own no-memory baseline. Measurement uncertainty is reported where the claims are made: the paired-contrast intervals of Figure~\ref{fig:contrasts} and the per-condition intervals of Appendix~\ref{full_tables_appendix}.}
  \label{fig:learning_curve}
\end{figure}

\section{Introduction}

Every interaction between an AI agent and its user or its environment is a learning opportunity, and today's agents miss almost all of them. An agent that resolves a difficult customer request today will start from zero when the same request arrives tomorrow. The underlying cause is architectural: agents are built on models whose weights are fixed at deployment, and only updated infrequently, if at all, during the lifetime of the system. Weight updates are the classical route to continual learning, but they carry heavy costs: training infrastructure, the risk of catastrophic forgetting, and a fresh evaluation and safety cycle for every update. For many organisations these costs are prohibitive and, in practice, a deployed agent is as capable on its last day as it was on its first.

In real-life deployments of AI agents there typically exists some form of supervision signal: did the customer accept the proposed solution, did the human worker need to edit the AI draft before sending it on, did the AI-constructed artefact pass the compliance checks. In this paper we investigate a mechanism that captures that signal for learning: pair the frozen model with an external memory. Feedback from each episode is distilled into compact natural-language insights, stored, and retrieved when they become relevant in later episodes. The system learns without making changes to the model weights, in a gradient-free continual learning cycle, running on supervision that is already available.

The industry's current answer to supplying agents with knowledge is retrieval-augmented generation (RAG), which offers a strong but static foundation. We use RAG as the baseline in this study.

We compare the baseline against learning from two grades of supervision, leading to the following three conditions:
\begin{itemize}
  \item \textbf{baseline}: the benchmark's stock agent, with BM25 retrieval over the complete corpus of roughly 700 policy documents that govern the tasks: a static RAG control.
  \item \textbf{experience}: the agent receives a single correct-or-incorrect verdict on the completed episode. It must generate its own lessons from that one bit: approaches that worked are kept, approaches that failed are marked to be avoided.
  \item \textbf{instruction}: a failed episode additionally receives the verified resolution, once, after the fact.
\end{itemize}

We measure this on the banking domain of $\tau$-bench~\cite{yao2024taubench, shi2026tauknowledge}, adapted in one respect: the memory store starts empty and stays active across a task's four trials, where standard stateless evaluation makes learning between trials impossible by construction. The baseline already retrieves over the complete policy corpus, so every improvement we measure is memory's contribution on top of static RAG, which is the question an enterprise deploying agents over its own document base faces. A second experiment freezes the stores built under instruction by two different models and has each read the other's, on tasks neither solves unaided: memory that transfers is what turns an agent's experience into organisational knowledge. The experiments use Spark, our shared memory system for AI agents (Section~\ref{spark_section}); an earlier study measured Spark in the coding domain~\cite{tablan2025smartertogether}, and this paper extends the evidence to a knowledge-intensive domain with no code involved.

The paper makes three contributions:

\textbf{Post-episode feedback is a sufficient learning signal.} From the one-bit verdict alone, the agent lifts pass\textasciicircum{}1 to 1.6 times the static-RAG baseline and holds banked solutions across trials at a rate of 0.88; with after-the-fact corrections it reaches 2.6 times the baseline and converts 22 of the 84 tasks the baseline never solves. The result replicates on a second, stronger model, where corrections lift pass\textasciicircum{}1 from 0.248 to 0.397.

\textbf{Accumulated memory transfers across models.} A frozen-store transfer experiment between Mistral Large and Claude Sonnet 5: each model, reading the store built by the other, rises above its own no-memory reference (pass\textasciicircum{}1 $+0.224$ for Mistral with the Sonnet-built store, $+0.066$ in the reverse direction, both resolved), and conditional on a task being transferable, conversion runs at roughly 31\% in both directions.

\textbf{A reusable measurement protocol.} Memory-active repeated trials on $\tau$-bench banking, with a static-RAG control and metrics that separate task success, reliability across trials, and knowledge retention, released with the harness and data.

\section{Related Work}
\label{related_work_section}

\textbf{Agent memory infrastructure.} A line of work builds storage and retrieval infrastructure for LLM memory. MemGPT manages the context window as an operating system manages virtual memory~\cite{packer2024memgptllmsoperatingsystems}; Mem0~\cite{chhikara2025mem0buildingproductionreadyai}, Zep~\cite{rasmussen2025zeptemporalknowledgegraph}, A-MEM~\cite{xu2025amemagenticmemoryllm}, MemOS~\cite{li2025memosmemoryosai}, and MemoryBank~\cite{Zhong_Guo_Gao_Ye_Wang_2024} provide long-term stores for conversational and personal-assistant settings, and several surveys map the space~\cite{zhang2024surveymemorymechanismlarge, wu2025humanmemoryaimemory, du2025rethinkingmemoryaitaxonomy}. Evaluation in this line centres on recall: benchmarks such as LongMemEval~\cite{wu2024longmemeval} and LoCoMo~\cite{maharana2024locomo} ask whether a fact from earlier in a long interaction can be recovered later. Guti\'errez et al.\ frame non-parametric memory as continual learning for LLMs~\cite{gutiérrez2025ragmemorynonparametriccontinual}, evaluated on retrieval and question answering. We ask a different question: whether memory changes the outcome of multi-step, tool-using tasks, measured under a reliability metric.

\textbf{Learning from experience with frozen weights.} Reflexion showed that an agent can improve within a task through verbal self-feedback~\cite{shinn2023reflexion}; Voyager grows a library of executable skills in an embodied setting~\cite{wang2023voyager}; ExpeL extracts reusable lessons from a training phase~\cite{zhao2023expel}; and Agent Workflow Memory induces workflows from successful trajectories~\cite{wang2024awm}. ReasoningBank is the closest prior system to our experience condition: it distils strategy-level lessons from both successes and failures, with an LLM judge supplying the verdict, reporting gains on web and coding benchmarks~\cite{ouyang2025reasoningbank}. Memento formalises case-based agent memory as a memory-augmented MDP~\cite{zhou2025mementofinetuningllmagents}. A recent line treats skill documents as the trainable external state of a frozen agent, the same framing we adopt for memory: Memento-Skills evolves executable skill folders~\cite{zhou2026mementoskills}, AutoSkill extracts and versions skills from interaction traces~\cite{yang2026autoskill}, Trace2Skill consolidates populations of labelled trajectories into a single skill document~\cite{ni2026trace2skill}, and SkillOpt optimises a skill document with the discipline of weight-space training~\cite{yang2026skillopt}. Across this line the agent is single-tenant: memory serves the agent that wrote it, verdicts come from LLM judges or held-out validation, and success is reported as single-trial rates. None separates learning from experience and learning from instruction as a measured axis, although Trace2Skill's ground-truth trajectory labels are a population-level form of instruction, and none measures reliability across repeated trials.

\textbf{Memory transfer across models.} MemCollab documents that naively transferring one model's memory to another can degrade performance below the no-memory baseline, because memory encodes the originating model's stylistic shortcuts alongside task knowledge, and responds with contrastive distillation of transferable invariants~\cite{chang2026memcollab}. Trace2Skill and SkillOpt both report that skills evolved with one model retain value for others~\cite{ni2026trace2skill, yang2026skillopt}. These transfer results are incidental to their designs; ours makes portability the measured object: frozen stores, read-only consumers, both directions, each consumer against its own no-memory baseline.

\textbf{Memory quality over time.} Zahn and Chana quantify how in-context memory erodes through its own lifecycle operations, losing 60\% of facts to compaction and silently dropping behavioural constraints~\cite{zahn2026facts}. AgingBench decomposes deployed-agent degradation into mechanisms, two of which, interference between similar entries and failure to propagate revised state, apply to any growing store~\cite{zhu2026agingbench}. Memora addresses the tension between abstraction and specificity in what a memory stores~\cite{memora2026}. These findings motivate two elements of our design: memory as a curated external store with an active lifecycle rather than accumulated context, and retention measured directly through hold and regression rates rather than assumed.

Relative to this literature, we ask whether the feedback ordinary operation already produces is a sufficient learning signal: two realistic feedback grades compared on the same tasks, on a hard, deterministic-reward benchmark where continual learning is directly observable. The control already retrieves over the complete task corpus, and portability is measured on frozen stores across models.

\section{Experimental Setup}
\label{setup_section}

\subsection{The Spark memory system}
\label{spark_section}

Spark is a shared memory service for AI agents~\cite{tablan2025smartertogether}. It runs as a standalone server that agents reach over the Model Context Protocol (MCP)~\cite{anthropic2024mcp}. The agent side holds no persistent state: everything the agent learns is written to the store, where it becomes available to every agent connected to the same memory, across sessions, users, and models. This subsection describes the parts of the system that are relevant for the experiments; the integration between Spark and the benchmark harness is described in Section~\ref{harness_section}.

\textbf{Insights.} The unit of knowledge is the \textbf{insight}: a compact natural-language statement of something worth remembering, such as a rule, a procedure, or a warning about an approach that fails. Each stored memory carries a title, its content, and a retrieval query, phrased as the question a future agent would ask mid-task; the query is the primary retrieval hook. Optional tags add soft relevance signals.

\textbf{The agent interface.} Agents interact with the store through four calls. \texttt{search} retrieves insights relevant to a query. \texttt{create\_memory} contributes a new insight. \texttt{enrich\_memory} adds information to an existing one. \texttt{share\_feedback} reports whether a retrieved insight proved useful, and accepts both positive and negative signals; this feedback feeds the trust estimate that, together with relevance, determines retrieval ranking. Together the four calls form a read-contribute-feedback cycle.

\textbf{Knowledge lifecycle.} Contributions are ingested asynchronously through an autonomous curation pipeline: a safety filter rejects malicious content, a quality gate rejects low-value content, and deduplication tests whether the contribution restates existing knowledge, in which case it is folded into the existing insight as supporting evidence. Beyond ingestion, curation operators run continuously against the store, pruning decayed insights and reconciling conflicts, and synthesis operators derive new insights that no agent contributed directly, by abstracting over clusters of related ones. In the current experimental setup, no step involves human curation.

\textbf{Domains.} A Spark server hosts multiple memory \textbf{domains}, each an isolated knowledge space with its own access scoping. The engine is shared across domains; domain specificity enters only through the vocabulary of the tool descriptions agents read and the prompts the internal operators use. Spark's default domain serves coding agents. The experiments in this paper use a \textbf{general-knowledge domain}, in which every description and prompt is domain-neutral, so the same configuration can serve any knowledge-intensive field.

\subsection{The $\tau$-Banking benchmark}
\label{benchmark_section}

Our apparatus builds on the banking domain of $\tau$-bench~\cite{yao2024taubench, barres2025tau2}, introduced by the $\tau$-Knowledge extension~\cite{shi2026tauknowledge}. Each task is a multi-turn conversation between the agent and a simulated customer, played by a separate LLM following a hidden persona and goal. The agent has domain tools that read from and write to a database, and a task succeeds when the final database state matches a pre-computed ground truth. Evaluation is programmatic and, for all but one of the 97 tasks, fully deterministic with no LLM judge involved. We grade under version 1.0.1 of the benchmark, which was current at the time of this study.

The banking domain withholds the policy. The agent receives no business rules upfront and must discover both the rules and the available tools by searching a corpus of roughly 700 interconnected policy documents with a retrieval tool (sparse BM25 retrieval in our configuration). This makes the domain hard: published frontier results sit around 25\% single-trial success, and even in the benchmark's gold configuration, which places the task-critical documents directly in the system prompt, the best models reach only about 40\%~\cite{shi2026tauknowledge}. Knowledge the model does not already hold is decisive, which is where memory can matter. This regime reflects most real-world agentic deployments: organisations own proprietary internal knowledge that was not part of the foundation model's training set.

The benchmark's primary metric is \textbf{pass\textasciicircum{}k}: a task counts only if the agent solves it in all $k$ independent trials. It is a deliberately strict reliability measure. Our protocol modifies its interpretation, as described below.

\textbf{Models.} We use two agent models, chosen to span the deployment spectrum: Mistral Large (version 2512), a capable open-weights model that an organisation with data sovereignty requirements could realistically self-host, and Claude Sonnet 5, a frontier model run at medium reasoning effort. The customer simulator is GPT-5.2, matching the benchmark's public leaderboard configuration.

\subsection{Experiment 1: continual learning}
\label{exp1_setup}

The first experiment measures continual learning from post-episode feedback. Both models run it on the full task set: Mistral Large under all three conditions, and Claude Sonnet 5 under the no-memory baseline and instruction, replicating the design at a higher capability level; the experience arm is omitted there on cost grounds. The definitions below apply identically to both models.

\subsubsection{Feedback signals}
\label{feedback_section}

Learning in our design is driven entirely by post-episode feedback. While a task runs, the agent has only its tools and the conversation: no ground truth of any form is available to it, and the task's hidden scenario is never available to the memory system. When the episode ends, feedback arrives in one of two grades.

The \textbf{outcome verdict} is a single bit: the final database state matches the ground truth or it does not. This is the weakest feedback a deployment produces; its real-world counterparts are a customer confirming resolution, a downstream check passing, or a human review signing off.

The \textbf{correction} is the task's verified action sequence, delivered once, after a failed episode. This is the feedback shape of a resolved ticket, an expert review, or a policy ruling: the right answer arrives from outside the agent, after the work is done.

The benchmark's deterministic evaluator instantiates both signals with perfect experimental reliability. It isolates the question of whether feedback of these shapes is sufficient for learning from the question of how noisy a given deployment's feedback is. We return to noise tolerance in Section~\ref{discussion_section}.

\subsubsection{Conditions}

We compare three conditions on the full task set. We call experience and instruction the two \textbf{learning modes}, after the feedback signal each learns from.

\textbf{No-memory baseline.} The stock benchmark agent. It retrieves over the complete policy corpus but retains nothing between trials; this is the static RAG control.

\textbf{Experience.} The memory agent, learning from the outcome verdict alone. The post-episode reflection receives the verdict and must distil its own lessons from the conversation and that single additional bit of information. The mode's limits are enforced in the write path: on a failed task the lesson is restricted to avoidance, recording what went wrong without asserting a correct action the agent cannot verify.

\textbf{Instruction.} The memory agent, learning from the verdict plus the correction. A failed episode's reflection additionally receives the task's verified action sequence; the correct knowledge is available to be written into memory exactly once, at the end of the failed episode.

\subsubsection{Memory integration}
\label{harness_section}

The memory agent extends the stock agent with four tools, \texttt{memory\_search}, \texttt{memory\_create}, \texttt{memory\_enrich}, and \texttt{memory\_feedback}, mapped one-to-one onto Spark's interface (Section~\ref{spark_section}). The tools are registered as first-class environment tools and execute through the benchmark's standard tool path, so memory calls receive no special handling in the conversation loop. The full harness is released with this paper\footnote{\url{https://github.com/memcoai/spark-continual-learning-paper-data}}, and the prompt fragments it adds to the stock agent are reproduced verbatim in Appendix~\ref{prompt_appendix}.

\textbf{Read path.} Retrieval is agent-driven: the harness never injects retrieved knowledge into the context, and every piece of remembered information reaches the agent through a \texttt{memory\_search} call the agent chose to make. A one-time reminder on the first customer turn tells the agent when to search, to search again when the situation changes mid-conversation, and to re-read any retrieved rule before committing to a consequential action; memory is presented as experience search, the sibling of the policy search, with the knowledge base authoritative on any conflict. Retrieval queries can draw only on the observable conversation. The reminder is reproduced in Appendix~\ref{prompt_appendix}.

\textbf{Write path.} All writes happen after the episode ends, in a dedicated reflection turn carried out by the same agent with the feedback in hand (Section~\ref{feedback_section}); writes attempted mid-conversation are acknowledged but not committed, which prevents the store from being poisoned by confident mistakes. What the reflection sees depends on the condition: under experience it has only the conversation and the verdict, and the verified solution is never delivered to any part of the system; under instruction, and only after a failed episode, the reflection additionally receives the verified action sequence. On instructed failures the reflection also receives a harness-computed diff of verified versus executed actions, so it locates the decisive divergence rather than guessing at it; on successes it receives the ordered list of consequential, state-changing actions the agent took, so the banked rule credits the decisive action rather than defaulting to the last one. The reflection operates under a contract that keeps the store precise: one rule in WHEN--THEN form per episode, decision-critical values verbatim and enforced by a write-time validator; explanations only where a tool result in the reflected conversation supports them; verified rules never contradicted or weakened, with one controlled merge for near-identical situations that prescribe different verified actions; and feedback submitted on the insights retrieved during the task. The full contract, with every prompt fragment, is reproduced in Appendix~\ref{prompt_appendix}.

\subsubsection{Protocol and metrics}
\label{protocol_section}

Each condition runs $k=4$ trials per task over the full set of 97 tasks, 388 simulated conversations per condition. In the memory conditions the store starts empty and remains active across a task's trials, so a lesson written after trial one can shape trial two; this is the property that makes continual learning measurable. Scheduling is trial-major: all tasks run their first trial before any task runs its second, which spaces a task's trials apart. Each condition uses its own fresh store, and the baseline touches no store.

We report pass\textasciicircum{}k computed exactly as the benchmark defines it, for comparability, with one interpretive caveat stated here once: under memory, trials are intentionally coupled, and rising consistency across trials is the phenomenon under study rather than a protocol violation. Alongside pass\textasciicircum{}k we report the \textbf{per-trial success curve}, the pooled success rate as a function of trial index, which a no-memory agent cannot improve by construction, and the \textbf{hold rate}, the probability that a task passed at trial $t$ passes again at $t+1$, with its complement measuring regressions. The hold rate is our operational measure of knowledge retention.

Pooled rates over trials are the primary statistics; per-task outcome grids are diagnostic and reported in Appendix~\ref{grids_appendix}. Trials within a task share difficulty and are correlated, so interval estimates treat the task as the sampling unit, using a cluster bootstrap; condition contrasts are paired per task. Appendix~\ref{stats_appendix} specifies the estimators and the resampling procedure. Because a large fraction of the task set is unsolved by the baseline in any trial, aggregate rates dilute the signal, so we additionally stratify tasks by baseline solvability. The \textbf{floor stratum} contains the tasks the baseline never solves in any of its four trials; the \textbf{competent stratum} contains the rest. A memory condition \textbf{converts} a floor task when it solves it in at least one trial. The stratum labels are fixed mechanically from the baseline arm's own results, before the memory arms are examined.

\subsection{Experiment 2: cross-model transfer}
\label{transfer_setup}

The second experiment tests whether the accumulated memory is portable across models: each model reads the store built by the other model's instruction run from Experiment~1, frozen at the end of the run. Nothing new is accumulated; each consumer is measured against its own Experiment~1 no-memory baseline. All cells cover the full task set; no run-time task selection occurs anywhere.

Consumers run read-only: search is available, writes and feedback are disabled, and the reflection is skipped, so the store is identical for every task and trial. Because the stores are frozen, trials are independent and there is no learning curve; we report the plain success rate over 4 trials per task.

For each consumer we define a transfer stratum: the tasks its own no-memory reference never solves in four trials, restricted to those the producer solved at least once during accumulation. This is where the foreign store demonstrably holds a working insight. The measured claims are two cross contrasts per consumer: one on the full task set, one on the transfer stratum. The two strata differ in size (49 and 7 tasks), and Appendix~\ref{transfer_appendix} gives the definitions and provenance.

\section{Results}
\label{results_section}

\subsection{Continual learning}
\label{exp1_results}

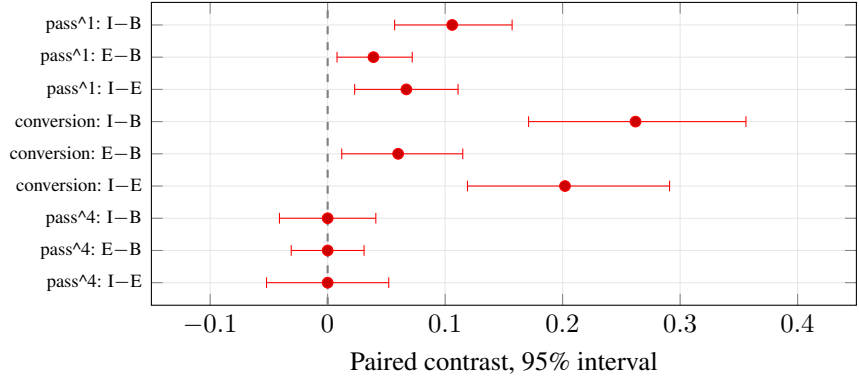
\begin{figure}[t]
  \centering
  \begin{tikzpicture}
    \begin{axis}[
        width=0.78\textwidth, height=0.40\textwidth,
        xlabel={Paired contrast, 95\% interval},
        ytick={1,...,9},
        yticklabels={
          {pass\textasciicircum{}4: I$-$E}, {pass\textasciicircum{}4: E$-$B}, {pass\textasciicircum{}4: I$-$B},
          {conversion: I$-$E}, {conversion: E$-$B}, {conversion: I$-$B},
          {pass\textasciicircum{}1: I$-$E}, {pass\textasciicircum{}1: E$-$B}, {pass\textasciicircum{}1: I$-$B}},
        yticklabel style={font=\scriptsize},
        ymin=0.3, ymax=9.7,
        xmin=-0.15, xmax=0.45,
        grid=major, grid style={line width=0.2pt, draw=gray!20},
      ]
      \addplot[dashed, gray, thick] coordinates {(0,0.3) (0,9.7)};
      \addplot+[only marks, mark=*, error bars/.cd, x dir=both, x explicit]
        coordinates {
          (0.106,9) += (0.051,0) -= (0.049,0)
          (0.039,8) += (0.033,0) -= (0.031,0)
          (0.067,7) += (0.044,0) -= (0.044,0)
          (0.262,6) += (0.094,0) -= (0.091,0)
          (0.060,5) += (0.055,0) -= (0.048,0)
          (0.202,4) += (0.089,0) -= (0.083,0)
          (0.000,3) += (0.041,0) -= (0.041,0)
          (0.000,2) += (0.031,0) -= (0.031,0)
          (0.000,1) += (0.052,0) -= (0.052,0)
        };
    \end{axis}
  \end{tikzpicture}
  \caption{The paired contrasts with 95\% cluster-bootstrap intervals. B: no-memory baseline; E: experience; I: instruction; conversion is the floor-stratum conversion rate. Positive values favour the first-named condition.}
  \label{fig:contrasts}
\end{figure}

Table~\ref{tab:threeway} reports every condition for both models; this section covers the three Mistral Large arms, and Section~\ref{sonnet_results} the replication. The ordering is the same at every $k$ below 4: instruction above experience above the no-memory baseline. At pass\textasciicircum{}4 the three conditions meet, for a structural reason taken up in Section~\ref{discussion_section}. On pass\textasciicircum{}1, instruction reaches 2.6 times the baseline and 1.6 times experience; the paired contrasts are $+0.106$ $[0.057, 0.157]$ over the baseline and $+0.067$ $[0.023, 0.111]$ over experience, with experience itself $+0.039$ $[0.008, 0.072]$ over the baseline (Figure~\ref{fig:contrasts}). The baseline sits at 0.064, well below the roughly 25\% frontier level on this benchmark. This is the regime the experiment targets: the outcome is decided by knowledge the model lacks.

\begin{table}[t]
  \caption{Point estimates on the full task set (97 tasks, 4 trials per task). Solved counts tasks passed at least once. \textbf{Hold}, the retention measure $P(\text{pass at } t{+}1 \mid \text{pass at } t)$, is reported only where trials are coupled through a changing store; for the baselines and the frozen-store consumers, whose trials are independent, it would measure re-draw persistence and is omitted. In the memory conditions pass\textasciicircum{}k reads differently from the stateless setting (Section~\ref{protocol_section}); only the baseline rows are comparable to published leaderboards. The transfer rows read the other model's frozen store (Section~\ref{transfer_setup}). Appendix~\ref{full_tables_appendix} gives 95\% intervals for every entry, including the omitted persistence values. Bold marks the best value per column within each continual-learning block.}
  \label{tab:threeway}
  \centering
  \small
  \begin{tabular}{llcccccc}
    \toprule
    Model & Condition & pass\textasciicircum{}1 & pass\textasciicircum{}2 & pass\textasciicircum{}3 & pass\textasciicircum{}4 & Solved & Hold \\
    \midrule
    \multicolumn{8}{l}{\emph{Continual learning}} \\
    Mistral Large & No-memory baseline & 0.064 & 0.038 & 0.034 & 0.031 & 13/97 & -- \\
    Mistral Large & Experience & 0.103 & 0.067 & 0.046 & 0.031 & 16/97 & \textbf{0.88} \\
    Mistral Large & Instruction & \textbf{0.170} & \textbf{0.091} & \textbf{0.057} & 0.031 & \textbf{32/97} & 0.65 \\
    \midrule
    Claude Sonnet 5 & No-memory baseline & 0.248 & 0.168 & 0.137 & 0.115 & 40/97 & -- \\
    Claude Sonnet 5 & Instruction & \textbf{0.397} & \textbf{0.268} & \textbf{0.206} & \textbf{0.165} & \textbf{62/97} & \textbf{0.83} \\
    \midrule\midrule
    \multicolumn{8}{l}{\emph{Cross-model transfer (frozen stores, read-only)}} \\
    Mistral Large & Reads Sonnet-built store & 0.289 & 0.172 & 0.124 & 0.093 & 51/97 & -- \\
    Claude Sonnet 5 & Reads Mistral-built store & 0.314 & 0.230 & 0.191 & 0.165 & 46/97 & -- \\
    \bottomrule
  \end{tabular}
\end{table}

\textbf{The learning curve.} The left panel of Figure~\ref{fig:learning_curve} shows the per-trial success rate. The three conditions are statistically indistinguishable on the cold first trial (6, 5, and 9 tasks of 97): an empty memory contributes nothing, so the conditions start as the same agent. The baseline stays flat across trials, which is re-draw noise from a stateless agent. Both memory conditions rise. Instruction climbs from 9 tasks at the cold trial to 18 on the second, regresses slightly on the third (16), and peaks at 23 on the fourth; experience rises monotonically to 14. The measured effect appears between trials, driven only by post-episode feedback.

\textbf{Stratification: where the gains come from.} The baseline solves 13 of 97 tasks in at least one trial, leaving 84 tasks it never solves. On the 84-task floor stratum, instruction converts 22 tasks, experience converts 5, and the baseline, by construction, converts none. The instruction-minus-experience conversion contrast is $+0.202$ $[0.119, 0.291]$, with instruction-minus-baseline at $+0.262$ $[0.171, 0.356]$. Ten of instruction's 22 conversions are tasks no other configuration ever solved. The conversion signature is visible in the per-task outcome grids (Appendix~\ref{grids_appendix}): a failed cold trial followed by warm-trial successes. This finding supports the designed mechanism: the verified rule is written into memory at the cold-trial failure, then retrieved and applied on the later trials. Experience's five conversions are of a different kind: with no verified solution available, it can only stabilise what the agent discovers through exploration guided away from previous failures.

\textbf{Retention.} The hold rates separate the two modes: experience holds banked solutions at 0.88, instruction at 0.65; the paired contrast is $-0.23$ $[-0.47, -0.02]$ (instruction minus experience).

\subsection{Replication on a second model}
\label{sonnet_results}

Claude Sonnet 5's two arms replicate the main result on a stronger model, running the full task set under the identical protocol.

The Sonnet rows of Table~\ref{tab:threeway} report both conditions. The baseline confirms the regime: Claude Sonnet 5 solves 40 of 97 tasks at least once, with pass\textasciicircum{}1 at 0.248 $[0.179, 0.323]$, consistent with the roughly 25\% frontier level quoted in Section~\ref{benchmark_section}, leaving a floor of 57 tasks it never solves. Instruction lifts every pass\textasciicircum{}k; the paired pass\textasciicircum{}1 contrast is $+0.149$ $[0.088, 0.211]$, larger in absolute terms than the $+0.106$ measured on Mistral Large. The lift survives, and grows, on a base model whose no-memory performance is four times higher, which addresses the concern that external memory only compensates for a weak model.

The learning-curve signature reproduces (Figure~\ref{fig:learning_curve}, centre panel): the baseline is flat across trials while instruction rises monotonically, from near-parity on the cold trial (26 tasks against the baseline's 22) to 48 of 97 by the fourth. On the 57-task floor stratum instruction converts 28 tasks, a conversion contrast of $+0.491$ $[0.362, 0.621]$ over a baseline that converts none by construction. The replication also resolves a contrast the Mistral experiment left directional: instruction's hold rate exceeds the baseline's re-draw persistence, 0.83 against 0.70 ($+0.126$ $[0.013, 0.261]$; Appendix~\ref{full_tables_appendix}).

\subsection{Cross-model transfer}
\label{exp2_results}

Both directions of transfer are resolved. Reading the frozen Sonnet-built store lifts Mistral Large from a pass\textasciicircum{}1 of 0.064 to 0.289 ($+0.224$ $[0.160, 0.289]$); reading the frozen Mistral-built store lifts Claude Sonnet 5 from 0.248 to 0.314 ($+0.066$ $[0.018, 0.119]$). Knowledge banked by one model helps the other, and the help runs uphill as well as down: the stronger model gains from rules the weaker one wrote. With frozen stores the consumer sits above its own baseline from the first trial; the transfer rows of Table~\ref{tab:threeway} and the right panel of Figure~\ref{fig:learning_curve} give the full picture.

Two readings stand out. First, transfer looks equally effective in both directions, even though the volumes differ. Conditional on a task being transferable, one the consumer never solves unaided but the producer solved during accumulation, conversion runs at roughly $31\%$ both ways: $+0.311$ $[0.222, 0.413]$ on the 49-task stratum, and $+0.321$ $[0.050, 0.625]$ on the 7-task one, the latter wide and read as directional. The full-set lifts differ by a factor of 3.4 only because the stronger producer had cleared far more of the other model's floor, 49 tasks against 7 (Appendix~\ref{transfer_appendix}): the directions differ in how much knowledge there is to move, and look alike in how effectively it moves.

Second, the results show the value of sharing memory across a diverse set of agents. Claude Sonnet 5 rises above its own baseline on rules banked by a model it outperforms four times over: the two models solve different tasks, so each store holds knowledge the other agent lacks. For the weaker model the effect is decisive: Mistral Large reading the Sonnet-built store outperforms its own instructed accumulation at every $k$ (0.289 against 0.170 on pass\textasciicircum{}1). And transferred knowledge stays once installed: the Sonnet consumer holds at 0.78, above its baseline's 0.70 re-draw persistence (Appendix~\ref{full_tables_appendix}).

\section{Discussion}
\label{discussion_section}

The three experiments support one claim from three directions. Feedback that deployments already produce, an outcome verdict or an after-the-fact correction, is enough for a frozen-weights agent to learn continually. The effect appears between trials, where nothing but the memory store carries state; it is replicated on models of significantly different baseline performance; and the accumulated store keeps its value when a different model reads it. On this benchmark, the gap between an agent that learns from its feedback and one that discards it is a factor of $2.6\times$ for Mistral Large and $1.6\times$ for Claude Sonnet 5 on single-trial success, both under instruction.

\textbf{Learning or answer caching?} The protocol invites an objection: the agent meets each task four times, so perhaps the store merely caches answers. Two properties of the results argue against that reading. The stored unit is a situational rule, a WHEN--THEN pair keyed to observable customer behaviour, and the task's hidden scenario never reaches the memory system, so a rule helps only if the agent recognises the live situation as one it applies to; each trial is a fresh conversation with a stochastic simulated customer, so what is measured is repetition with variation. And the stores transfer: naively transferred memory often encodes the producing model's stylistic shortcuts and can push a consumer below its own baseline~\cite{chang2026memcollab}, while here both directions lift the consumer, so the stores hold task knowledge stated generally enough for a different model to apply. What the design does not measure is generalisation to unseen task types; recurring situations are the deployment case this study targets, and the earlier coding-domain study~\cite{tablan2025smartertogether} carries the evidence for breadth.

\textbf{Does memory disturb what the agent already solves?} A store of learned rules could mislead the agent on tasks it handles well unaided, and the competent stratum measures this: of the 13 tasks the Mistral Large baseline solves, experience retains 11 and instruction 10, losses within what re-draw noise alone would produce, while tasks solved at least once rise from 13 to 16 and 32 (Table~\ref{tab:threeway}). We believe the occasional misapplication is itself circumstantial evidence against caching: a cached answer is inert on tasks it does not match, while a rule general enough to transfer across models is also general enough to misfire on a look-alike situation.

\textbf{What pass\textasciicircum{}k measures under memory.} pass\textasciicircum{}k is the benchmark's reliability metric: a task counts only if the agent solves it in every one of $k$ trials, estimated as the probability that $k$ of the task's four recorded trials, drawn at random, all pass. On pass\textasciicircum{}4 the three Mistral Large conditions are identical, all at 0.031. Yet the conditions differ substantially: instruction converts 22 tasks the baseline never solves, and experience holds $88\%$ of its banked solutions. The explanation is structural, and one example grid makes it concrete: a task converted after a failed first trial and held thereafter, $0111$, contributes 0.75 at pass\textasciicircum{}1, 0.5 at pass\textasciicircum{}2, 0.25 at pass\textasciicircum{}3, and exactly 0 at pass\textasciicircum{}4 under the benchmark's estimator (Appendix~\ref{stats_appendix}). At pass\textasciicircum{}4 the metric cannot distinguish that task from one never solved, so a learning agent's advantage should shrink as $k$ grows and vanish at $k=4$. The data trace that profile: instruction's lead over the baseline runs 0.106, 0.053, 0.023, 0.000 across the four $k$ (Table~\ref{tab:threeway}), and the paired pass\textasciicircum{}4 contrasts sit at zero with intervals spanning it (Figure~\ref{fig:contrasts}). For this reason we use the hold rate as the measure of consistency once knowledge is in the store, and report pass\textasciicircum{}k for comparability with the benchmark's convention; the protocol, with the per-trial curve for acquisition and the hold rate for retention, is released for reuse.

\textbf{From reliable feedback to deployment feedback.} The evaluator delivers both feedback signals with perfect reliability; deployment signals arrive late, missing, or wrong (Section~\ref{feedback_section}). Feedback noise is outside this paper's scope: it is the concern of other Spark functionality, including trust-weighted retrieval and autonomous curation.

\textbf{Limitations.} Every condition is a single run, so the reported intervals speak to task sampling and not to run-to-run variation. The component this leaves out, cross-task coupling through the shared store, appears weak in this experiment: under trial-major scheduling a late task's first trial already runs against a store holding most earlier tasks' lessons, yet first-trial rates sit at baseline parity in both models, so the run's randomness is largely per-task and travels with the resampled task. We hypothesise this reflects the benchmark's design, in which tasks are largely independent scenarios that each turn on their own decisive rule; deployments where lessons transfer across tasks would need repeated runs to estimate that variance. The experience mode ran only on Mistral Large; Claude Sonnet 5 ran the baseline and instruction. On a stronger model, where first solves are more common and experience has more successes to bank, we expect it to acquire more while keeping its high retention; the design can test this directly. Beyond that, the evidence covers two models and one domain, with transfer measured from instruction-built stores.

\section{Conclusion}

Deployed agents produce feedback in the ordinary course of their work, and this paper shows that the feedback is enough to learn from. Paired with an external memory, a frozen-weights agent turned outcome verdicts and after-the-fact corrections into lasting capability: 2.6 times the static-RAG baseline on single-trial success for Mistral Large, a replicated lift on Claude Sonnet 5, and 22 of 84 previously unsolved tasks converted, all without touching a model weight. The accumulated stores carried their value across models, in both directions, which turns one agent's experience into knowledge an organisation can hand to any of its agents. The harness, protocol, and data are released for others to build on.

\clearpage

\bibliographystyle{ieeetr}
{\small
\bibliography{memco_continual_learning}

\begin{thebibliography}{10}

\bibitem{yao2024taubench}
S.~Yao, N.~Shinn, P.~Razavi, and K.~Narasimhan, ``$\tau$-bench: A benchmark for tool-agent-user interaction in real-world domains,'' 2024.

\bibitem{shi2026tauknowledge}
Q.~Shi, A.~Zytek, P.~Razavi, K.~Narasimhan, and V.~Barres, ``$\tau$-knowledge: Evaluating conversational agents over unstructured knowledge,'' 2026.

\bibitem{tablan2025smartertogether}
V.~Tablan, S.~Taylor, G.~Hurtado, K.~Bernhem, A.~Uhrenholt, G.~Farei, and K.~Moilanen, ``Smarter together: Creating agentic communities of practice through shared experiential learning,'' 2025.

\bibitem{packer2024memgptllmsoperatingsystems}
C.~Packer, S.~Wooders, K.~Lin, V.~Fang, S.~G. Patil, I.~Stoica, and J.~E. Gonzalez, ``Memgpt: Towards llms as operating systems,'' 2024.

\bibitem{chhikara2025mem0buildingproductionreadyai}
P.~Chhikara, D.~Khant, S.~Aryan, T.~Singh, and D.~Yadav, ``Mem0: Building production-ready ai agents with scalable long-term memory,'' 2025.

\bibitem{rasmussen2025zeptemporalknowledgegraph}
P.~Rasmussen, P.~Paliychuk, T.~Beauvais, J.~Ryan, and D.~Chalef, ``Zep: A temporal knowledge graph architecture for agent memory,'' 2025.

\bibitem{xu2025amemagenticmemoryllm}
W.~Xu, Z.~Liang, K.~Mei, H.~Gao, J.~Tan, and Y.~Zhang, ``A-mem: Agentic memory for llm agents,'' 2025.

\bibitem{li2025memosmemoryosai}
Z.~Li, S.~Song, C.~Xi, H.~Wang, C.~Tang, S.~Niu, D.~Chen, J.~Yang, C.~Li, Q.~Yu, J.~Zhao, Y.~Wang, P.~Liu, Z.~Lin, P.~Wang, J.~Huo, T.~Chen, K.~Chen, K.~Li, Z.~Tao, H.~Lai, H.~Wu, B.~Tang, Z.~Wang, Z.~Fan, N.~Zhang, L.~Zhang, J.~Yan, M.~Yang, T.~Xu, W.~Xu, H.~Chen, H.~Wang, H.~Yang, W.~Zhang, Z.-Q.~J. Xu, S.~Chen, and F.~Xiong, ``Memos: A memory os for ai system,'' 2025.

\bibitem{Zhong_Guo_Gao_Ye_Wang_2024}
W.~Zhong, L.~Guo, Q.~Gao, H.~Ye, and Y.~Wang, ``Memorybank: Enhancing large language models with long-term memory,'' {\em Proceedings of the AAAI Conference on Artificial Intelligence}, vol.~38, pp.~19724--19731, Mar. 2024.

\bibitem{zhang2024surveymemorymechanismlarge}
Z.~Zhang, X.~Bo, C.~Ma, R.~Li, X.~Chen, Q.~Dai, J.~Zhu, Z.~Dong, and J.-R. Wen, ``A survey on the memory mechanism of large language model based agents,'' 2024.

\bibitem{wu2025humanmemoryaimemory}
Y.~Wu, S.~Liang, C.~Zhang, Y.~Wang, Y.~Zhang, H.~Guo, R.~Tang, and Y.~Liu, ``From human memory to ai memory: A survey on memory mechanisms in the era of llms,'' 2025.

\bibitem{du2025rethinkingmemoryaitaxonomy}
Y.~Du, W.~Huang, D.~Zheng, Z.~Wang, S.~Montella, M.~Lapata, K.-F. Wong, and J.~Z. Pan, ``Rethinking memory in ai: Taxonomy, operations, topics, and future directions,'' 2025.

\bibitem{wu2024longmemeval}
D.~Wu, H.~Wang, W.~Yu, Y.~Zhang, K.-W. Chang, and D.~Yu, ``Longmemeval: Benchmarking chat assistants on long-term interactive memory,'' 2024.

\bibitem{maharana2024locomo}
A.~Maharana, D.-H. Lee, S.~Tulyakov, M.~Bansal, F.~Barbieri, and Y.~Fang, ``Evaluating very long-term conversational memory of llm agents,'' 2024.

\bibitem{gutiérrez2025ragmemorynonparametriccontinual}
B.~J. Gutiérrez, Y.~Shu, W.~Qi, S.~Zhou, and Y.~Su, ``From rag to memory: Non-parametric continual learning for large language models,'' 2025.

\bibitem{shinn2023reflexion}
N.~Shinn, F.~Cassano, E.~Berman, A.~Gopinath, K.~Narasimhan, and S.~Yao, ``Reflexion: Language agents with verbal reinforcement learning,'' 2023.

\bibitem{wang2023voyager}
G.~Wang, Y.~Xie, Y.~Jiang, A.~Mandlekar, C.~Xiao, Y.~Zhu, L.~Fan, and A.~Anandkumar, ``Voyager: An open-ended embodied agent with large language models,'' 2023.

\bibitem{zhao2023expel}
A.~Zhao, D.~Huang, Q.~Xu, M.~Lin, Y.-J. Liu, and G.~Huang, ``Expel: Llm agents are experiential learners,'' 2023.

\bibitem{wang2024awm}
Z.~Z. Wang, J.~Mao, D.~Fried, and G.~Neubig, ``Agent workflow memory,'' 2024.

\bibitem{ouyang2025reasoningbank}
S.~Ouyang, J.~Yan, I.-H. Hsu, Y.~Chen, K.~Jiang, Z.~Wang, R.~Han, L.~T. Le, S.~Daruki, X.~Tang, V.~Tirumalashetty, G.~Lee, M.~Rofouei, H.~Lin, J.~Han, C.-Y. Lee, and T.~Pfister, ``Reasoningbank: Scaling agent self-evolving with reasoning memory,'' 2025.

\bibitem{zhou2025mementofinetuningllmagents}
H.~Zhou, Y.~Chen, S.~Guo, X.~Yan, K.~H. Lee, Z.~Wang, K.~Y. Lee, G.~Zhang, K.~Shao, L.~Yang, and J.~Wang, ``Memento: Fine-tuning llm agents without fine-tuning llms,'' 2025.

\bibitem{zhou2026mementoskills}
H.~Zhou, S.~Guo, A.~Liu, Z.~Yu, Z.~Gong, B.~Zhao, Z.~Chen, M.~Zhang, Y.~Chen, J.~Li, R.~Yang, Q.~Liu, X.~Yu, J.~Zhou, N.~Wang, C.~Sun, and J.~Wang, ``Memento-skills: Let agents design agents,'' 2026.

\bibitem{yang2026autoskill}
Y.~Yang, J.~Li, Q.~Pan, B.~Zhan, Y.~Cai, L.~Du, J.~Zhou, K.~Chen, Q.~Chen, X.~Li, B.~Zhang, and L.~He, ``Autoskill: Experience-driven lifelong learning via skill self-evolution,'' 2026.

\bibitem{ni2026trace2skill}
J.~Ni, Y.~Liu, X.~Liu, Y.~Sun, M.~Zhou, P.~Cheng, D.~Wang, E.~Zhao, X.~Jiang, and G.~Jiang, ``Trace2skill: Distill trajectory-local lessons into transferable agent skills,'' 2026.

\bibitem{yang2026skillopt}
Y.~Yang, Z.~Gong, W.~Huang, Q.~Yang, Z.~Zhou, Z.~Huang, Y.~Li, X.~Gao, Q.~Dai, B.~Liu, K.~Qiu, Y.~Yang, D.~Chen, X.~Yang, and C.~Luo, ``Skillopt: Executive strategy for self-evolving agent skills,'' 2026.

\bibitem{chang2026memcollab}
Y.~Chang, Y.~Wu, Q.~Wu, and L.~Lin, ``Memcollab: Cross-agent memory collaboration via contrastive trajectory distillation,'' 2026.

\bibitem{zahn2026facts}
O.~Zahn and S.~Chana, ``Facts as first-class objects: Knowledge objects for persistent llm memory,'' 2026.

\bibitem{zhu2026agingbench}
J.~Zhu, Y.~Ro, J.~Robertson, K.~Wang, J.~Li, H.~Vikalo, A.~Akella, and Z.~Wang, ``Your agents are aging too: Agent lifespan engineering for deployed systems,'' 2026.

\bibitem{memora2026}
M.~Xia, X.~Zhang, S.~Dixit, P.~Harimurugan, R.~Wang, V.~Ruhle, R.~Sim, C.~Bansal, and S.~Rajmohan, ``Memora: A harmonic memory representation balancing abstraction and specificity,'' 2026.

\bibitem{anthropic2024mcp}
{Anthropic}, ``Model context protocol,'' 2024.
\newblock Open protocol specification.

\bibitem{barres2025tau2}
V.~Barres, H.~Dong, S.~Ray, X.~Si, and K.~Narasimhan, ``$\tau^2$-bench: Evaluating conversational agents in a dual-control environment,'' 2025.

\end{thebibliography}
}

\clearpage
\appendix

\section{Harness prompts}
\label{prompt_appendix}

This appendix reproduces, verbatim, the prompt fragments the harness adds to the stock benchmark agent (Section~\ref{harness_section}). Text in curly braces marks placeholders the harness fills at run time.

\subsection{Memory advertisement in the system prompt}

Inserted into the domain policy immediately after its existing instruction to use the knowledge-base search tool, so the two retrieval channels are advertised side by side and memory remains subordinate to the policy and knowledge base:

\begin{lstlisting}
## Shared experience memory

**Search your experience memory** using the provided `memory_search` tool before completing a task. This contains knowledge learned in the course of applying policies, and lessons from past similar interactions.
\end{lstlisting}

\subsection{Memory tool descriptions}

The four memory tools are described to the agent in the same style as the domain's knowledge-base search tool. The tool schemas, as the agent sees them:

\begin{lstlisting}
memory_search(query, tags?)
  Search your memory of past customer interactions for relevant experiences
  (what worked or didn't go well in the past).
  Use it whenever you are starting to handle a customer request, or before
  taking an important action -- to reuse what worked before and avoid known
  pitfalls.
  Args:
    query: What you want to recall, phrased as a question or situation
      (e.g. "highest cash back card with no annual fee for frequent traveller").
    tags: Optional context tags to focus the search.

memory_create(query, title, content, tags?)
  Save a new lesson to memory so future agents handle similar requests better.
  Record one concise, reusable takeaway from this interaction -- a rule on how
  to apply a policy, the correct action for a situation, or a pitfall to avoid.
  Capture the generalizable pattern (customer need -> right action), not
  one-off facts like names or account numbers.
  Args:
    query: What a future agent would search to find this lesson.
    title: A short title.
    content: The concise, reusable lesson.
    tags: Optional context tags.

memory_enrich(memory_idx, title, content, session_id, tags?)
  Extend a remembered lesson you retrieved with what you just learned.
  When a memory from your search was close but incomplete or needs correcting,
  enrich it instead of creating a duplicate. Requires the memory_idx and
  session_id from a search you ran earlier in this conversation.

memory_feedback(feedback, session_id)
  Rate the remembered insights you were shown, to improve future recall.
  After you act, tell memory which retrieved insights were relevant and
  correct (or not), so good lessons surface more and bad ones less. Requires
  the session_id from the search that returned them.
\end{lstlisting}

\subsection{The memory reminder}

Appended once, to the first customer message of each conversation:

\begin{lstlisting}
<memory_check>
After you've collected all the data, and before you respond to the customer, check your experience memory. Call the `memory_search` tool with a query describing this customer's request and your intended solution to check if you need to change anything before responding.
If the situation changes later in the conversation (a new procedure, code, or request appears), call `memory_search` again with the new specifics before acting on it.
Before you commit to a recommendation or perform a consequential account action, re-read any VERIFIED RULE from your memory search results: follow it, or state explicitly why it does not apply to this customer before deviating.
</memory_check>
\end{lstlisting}

\subsection{The reflection turn}
\label{reflection_prompt_appendix}

The post-task reflection is delivered as a user message built from the template below; the completed conversation is seeded as real message history, so the template carries only the new information (verdict, correction, diffs) and the instructions.

\begin{lstlisting}
The customer interaction is now COMPLETE.

{verdict}
{correction_block}{diff_block}{consequential_block}{outcome_block}{diagnosis_hint}
Now update your experience memory so future similar interactions go better. This is the ONLY thing to do now -- do NOT message the customer and do NOT call any other tools (no KB_search, no apply_for_credit_card):

{save_instructions}

Rules for EVERY memory you write:
- NEVER contradict, weaken, or add exceptions to an existing memory whose title starts with "VERIFIED RULE:". If your experience seems to conflict with one, save a SEPARATE note describing the conflict -- do not edit the rule. (One exception: merging this task's verified action with a retrieved VERIFIED RULE for a look-alike situation into a single branch rule, where the save instructions above permit it -- both verified actions are kept.)
- Do NOT state facts (rates, fees, eligibility, program terms) that no tool result in THIS conversation supports. If you cannot point to the evidence for a reason, write "Reason not documented in the KB; rule verified against the ground-truth outcome." instead of a reason.
- Do not save customer-specific facts (names, account numbers).

{rate_line}{transcript_block}
\end{lstlisting}

Under the deterministic evaluator, \texttt{\{verdict\}} is rendered as:

\begin{lstlisting}
TASK OUTCOME (ground truth): the desired outcome WAS / was NOT achieved.
\end{lstlisting}

On any failure, \texttt{\{diagnosis\_hint\}} adds:

\begin{lstlisting}
A wrong outcome doesn't always mean wrong facts. The cause may be the wrong action or procedure, a missing or mis-stated detail, the wrong tool -- or, when the customer is the one who must act, not persuading them to take the right step. Work out which applied here and capture that lesson.
\end{lstlisting}

\subsection{First-lesson instructions}

The \texttt{\{save\_instructions\}} block is selected by outcome and learning mode. On success (both modes), when the harness-computed list of consequential actions is available:

\begin{lstlisting}
**1. Bank the win (required -- do this FIRST).** Call `memory_create` with the ONE decisive rule that made this task succeed:
- title: `VERIFIED RULE: <short name for this customer situation>`
- content: one sentence of the form "WHEN <the customer situation, as it appeared in this conversation> THEN <the DECISIVE action(s), chosen from the CONSEQUENTIAL ACTIONS list above -- tool names and every decision-critical value (product/card names, account classes, reasons, amounts) verbatim; customer identifiers as placeholders like customer_name=<their name>>".
- The decisive action is the one that produced the graded outcome. It is often NOT the last thing that happened: a closing transfer, a goodbye, or an escalation at the end of the conversation is rarely what made the task succeed. Pick from the list above the action(s) that actually resolved the customer's request.
- The WHEN clause must name the customer's goal AND the observable behavior that distinguishes this situation from similar-looking ones (what the customer insisted on, accepted, or pivoted to in THIS conversation). Describe only behavior that actually occurred.
- Record the rule that actually WON. Do NOT soften it with alternatives or conditions you did not verify this time ("other options may be better if...") -- hedged rules do not survive the next conversation.

**2. Optionally save further lessons** -- a separate `memory_create` per genuinely DISTINCT, reusable lesson. Prefer a few high-value lessons over many marginal ones.
\end{lstlisting}

A variant differing only in not referencing the consequential-actions list is used when that list is empty. On failure under \textbf{instruction}, with the verified solution and action diff in context:

\begin{lstlisting}
**1. Save the verified rule (required -- do this FIRST).** Call `memory_create` with ONE decisive rule that would have made this task succeed, built from the CORRECT solution above:
- title: `VERIFIED RULE: <short name for this customer situation>`
- content: one sentence of the form "WHEN <the customer situation, as it appeared in this conversation> THEN <the exact correct action(s), copied VERBATIM from the CORRECT solution above -- tool names and every decision-critical value (product/card names, account classes, reasons, amounts, timestamps) unchanged>". Only customer identifiers (name, income, ids) become placeholders like customer_name=<their name>. If the ACTION DIFF above shows the decisive divergence, add one sentence naming it (the extra action to avoid, or the exact argument to get right).
- The WHEN clause must name the customer's goal AND the observable behavior that distinguishes this situation from similar-looking ones (what the customer insisted on, accepted, or pivoted to in THIS conversation) -- so a future agent facing a look-alike can tell whether this rule applies. Describe only behavior that actually occurred.
- Do NOT soften the rule with conditions, alternatives, or exceptions that are not in the CORRECT solution ("if the customer prefers X, then Y may be better" is how future agents fail). The rule is verified by the ground truth: state it as a directive.

**2. Optionally save further lessons** -- a separate `memory_create` per genuinely DISTINCT, reusable lesson (e.g. a procedural pitfall). Prefer a few high-value lessons over many marginal ones.
\end{lstlisting}

On failure under \textbf{experience}, with no verified solution available, the lesson is restricted to avoidance:

\begin{lstlisting}
**1. Save the warning (required -- do this FIRST).** Call `memory_create` recording what to AVOID, from what actually happened:
- title: `AVOID: <the mistaken action, short>`
- content: one sentence of the form "WHEN <the customer situation, as it appeared in this conversation> do NOT <the exact action(s) taken that led to the failed outcome -- tool names and decision-critical values (product/card names, account classes, amounts) verbatim; customer identifiers as placeholders> -- verified wrong against the ground-truth outcome."
- Name the concrete action from THIS conversation, not a vague behavior ("be more careful").
- The WHEN clause must name the customer's goal AND the observable behavior that distinguishes this situation from similar-looking ones (what the customer insisted on, accepted, or pivoted to in THIS conversation). Describe only behavior that actually occurred.
- You do NOT know what the correct answer was: do not invent one, and do not guess.

**2. Optionally save further lessons** -- a separate `memory_create` per genuinely DISTINCT, reusable lesson. Prefer a few high-value lessons over many marginal ones.
\end{lstlisting}

The success and instruction-failure instructions are followed by the twin-situation guidance that governs the single permitted merge of verified rules:

\begin{lstlisting}
**Twin situations.** If a retrieved memory titled "VERIFIED RULE:" covers a situation that LOOKS like this one but prescribes a DIFFERENT action than this task's verified action, do NOT write a second competing rule -- write ONE branch rule as your step-1 memory instead:
- title: `VERIFIED RULE: <the shared situation> -- which action applies`
- content: "WHEN <the shared situation>: IF <what the customer observably said or did in THIS conversation> THEN <this task's verified action>; IF <the customer behavior from the retrieved rule's situation> THEN <the retrieved rule's action, unchanged>." End with "Supersedes: <title of the retrieved rule>".
- Each branch's action must come VERBATIM from its verified source (this task's verified action, or the retrieved VERIFIED RULE) -- never invent a branch or a condition. Every branch condition must be something a future agent can OBSERVE the customer say or do.
- Combining two verified rules this way is the ONE permitted exception to the never-contradict-a-VERIFIED-RULE rule: both verified actions are preserved.
\end{lstlisting}

\subsection{Feedback rating rubrics}

The \texttt{\{rate\_line\}} instruction asks the agent to rate the insights retrieved during the task, and appends an outcome-anchored rubric to prevent the ratings from following the agent's opinion of its own performance rather than the ground truth. By outcome:

\begin{lstlisting}
Success: Rating rules: insights that pointed toward the action that succeeded are correct=true. Rate the rest on factual accuracy only; do NOT rate a "VERIFIED RULE:" insight incorrect merely because it was not needed this time.

Failure (instruction): Rating rules: an insight is correct=true ONLY if it points toward the CORRECT solution above. An insight that endorsed what you actually did this time is correct=false -- the ground truth proves it. Never rate a "VERIFIED RULE:" insight incorrect for being too strict.

Failure (experience): Rating rules: insights that endorsed the approach you took are correct=false -- the ground truth proves it. Do not rate an insight correct because it sounded authoritative. Never rate a "VERIFIED RULE:" insight incorrect for being too strict.
\end{lstlisting}

\subsection{Write-time validation}

When the pinned rule omits decision-critical strings from the verified solution or the action diff, the write is refused and the tool result returns the exact gaps, for a bounded number of retries:

\begin{lstlisting}
NOT SAVED -- the VERIFIED RULE is missing decision-critical content from the CORRECT solution / ACTION DIFF: {missing strings}. Each of these must appear VERBATIM in the rule (in the THEN clause as a directive -- not as a placeholder like <current_time>, and not inside an 'e.g.' example). Call memory_create again now with the corrected content.
\end{lstlisting}

\section{Transfer experiment details}
\label{transfer_appendix}

\paragraph{Strata.} Every run covers the full 97-task set; the strata are analysis-time objects, fixed from the reference and producer runs and committed before either consumer cell started. For each consumer, its \textbf{floor} is the set of tasks its no-memory reference never solves in four trials: 84 tasks for Mistral Large, 57 for Claude Sonnet 5, the latter a strict subset of the former. A store's \textbf{coverage} is the set of tasks its producer solved at least once during instructed accumulation: 32 tasks for the Mistral store, 62 for the Sonnet store. A consumer's transfer stratum is its floor intersected with the foreign store's coverage: 49 tasks for the Sonnet-to-Mistral direction, 7 for Mistral-to-Sonnet. The asymmetry is a measurement about the models, made before the transfer cells ran: Sonnet's floor sits almost entirely inside the harder region of the task set, where Mistral's instructed run converted little, so the Mistral-to-Sonnet stratum contrast has little power and is reported descriptively, with the full-set contrast carrying that direction.

\paragraph{Store provenance.} Each store is frozen at the end of its producer's instructed run: the Experiment 1 instruction arm for Mistral Large, and the Section~\ref{sonnet_results} instruction arm for Claude Sonnet 5. Consumer cells authenticate with read-only credentials, the harness registers only the search tool, and store integrity (insight count and latest write timestamp) is verified before and after each cell.

\clearpage
\section{Per-task outcome grids}
\label{grids_appendix}

Tables~\ref{tab:grids_a} and~\ref{tab:grids_b} list the outcome grid of every task: four characters in trial order, with 1 for a passed trial and 0 for a failed one; a dash marks the single simulation lost to an infrastructure error. Task identifiers are the benchmark's own: the 97 released tasks carry non-contiguous identifiers running from 001 to 102, and all of them are included here. The grids are diagnostic; the pooled statistics of the main text remain the reported measures.

\begin{center}
  \captionof{table}{Per-task outcome grids, tasks 001--054. Columns follow the blocks of Table~\ref{tab:threeway}: the four Mistral Large runs, then the three Claude Sonnet 5 runs; \emph{S store} and \emph{M store} are the frozen-store consumer cells.}
  \label{tab:grids_a}
  \scriptsize\ttfamily
  \begin{tabular}{lccccccc}
    \toprule
    & \multicolumn{4}{c}{\rmfamily Mistral Large} & \multicolumn{3}{c}{\rmfamily Claude Sonnet 5} \\
    \cmidrule(lr){2-5} \cmidrule(lr){6-8}
    \rmfamily Task & \rmfamily Base & \rmfamily Exp & \rmfamily Instr & \rmfamily S store & \rmfamily Base & \rmfamily Instr & \rmfamily M store \\
    \midrule
    001 & 0010 & 0111 & 1111 & 1011 & 1111 & 1111 & 1111 \\
    002 & 0000 & 0111 & 0011 & 1111 & 1111 & 1111 & 1101 \\
    003 & 0000 & 0000 & 0000 & 1001 & 0000 & 0111 & 0000 \\
    004 & 0000 & 0001 & 0101 & 1111 & 1111 & 1111 & 1111 \\
    005 & 0010 & 0001 & 0001 & 0101 & 1000 & 0011 & 1101 \\
    006 & 1101 & 0011 & 1101 & 1111 & 1111 & 1111 & 1111 \\
    007 & 1111 & 1111 & 0111 & 0111 & 0111 & 0001 & 1111 \\
    008 & 1000 & 0011 & 0111 & 1110 & 1111 & 1111 & 0111 \\
    010 & 1111 & 1110 & 1111 & 0011 & 0001 & 1000 & 0110 \\
    012 & 0000 & 0000 & 0111 & 1111 & 0110 & 1111 & 1111 \\
    014 & 0000 & 0000 & 0000 & 0001 & 0101 & 0011 & 0000 \\
    015 & 0000 & 0000 & 0001 & 1111 & 0000 & 0000 & 0000 \\
    016 & 0000 & 0011 & 0001 & 0001 & 1111 & 1111 & 1111 \\
    017 & 0000 & 0000 & 0001 & 1000 & 1111 & 1111 & 0111 \\
    018 & 0000 & 0000 & 0000 & 0000 & 0101 & 0111 & 0101 \\
    019 & 0000 & 0000 & 1000 & 0011 & 1011 & 1100 & 1111 \\
    020 & 0000 & 0000 & 0000 & 0000 & 0000 & 0001 & 1000 \\
    021 & 0000 & 0000 & 0000 & 0000 & 0111 & 0111 & 1111 \\
    022 & 0000 & 0000 & 0000 & 1000 & 0000 & 0101 & 0000 \\
    023 & 0001 & 0000 & 0000 & 0000 & 0010 & 1000 & 1000 \\
    024 & 0000 & 0000 & 0111 & 1111 & 0000 & 0001 & 0011 \\
    025 & 0100 & 0111 & 0111 & 1110 & 0010 & 1110 & 0000 \\
    026 & 0000 & 0000 & 0000 & 0001 & 0000 & 0000 & 0000 \\
    027 & 0000 & 0000 & 0000 & 0000 & 0000 & 0010 & 0000 \\
    028 & 0000 & 0000 & 0000 & 0001 & 0000 & 1000 & 0001 \\
    029 & 0000 & 0000 & 0000 & 0000 & 0000 & 0110 & 0000 \\
    031 & 0000 & 0000 & 0111 & 1001 & 1110 & 1111 & 1111 \\
    032 & 0000 & 0001 & 1100 & 1110 & 1111 & 1111 & 1111 \\
    033 & 0001 & 1011 & 1011 & 1110 & 1101 & 1010 & 0111 \\
    034 & 1111 & 1111 & 1010 & 1110 & 0011 & 1100 & 0101 \\
    035 & 0001 & 1111 & 1101 & 1111 & 1111 & 1111 & 1111 \\
    036 & 0100 & 0000 & 0000 & 0010 & 1110 & 0111 & 0100 \\
    037 & 0000 & 0000 & 0000 & 0010 & 0111 & 1111 & 0111 \\
    038 & 0000 & 0000 & 0010 & 0000 & 0001 & 1100 & 0110 \\
    039 & 0000 & 0000 & 0000 & 0000 & 0000 & 0000 & 0000 \\
    040 & 0000 & 0000 & 0000 & 0000 & 0000 & 0100 & 0000 \\
    041 & 0000 & 0000 & 0000 & 0000 & 0000 & 0000 & 0000 \\
    043 & 0000 & 0000 & 0000 & 0000 & 0000 & 0111 & 0000 \\
    044 & 0000 & 0000 & 0000 & 0000 & 0000 & 0000 & 0010 \\
    045 & 0000 & 0000 & 0000 & 0000 & 0000 & 1111 & 0000 \\
    046 & 0000 & 0000 & 0000 & 0000 & 0000 & 0111 & 0000 \\
    047 & 0000 & 0000 & 0000 & 1000 & 0000 & 0010 & 0000 \\
    048 & 0000 & 0000 & 0000 & 0000 & 0000 & 0000 & 0000 \\
    049 & 0000 & 0000 & 0000 & 0000 & 0000 & 0000 & 0000 \\
    050 & 0000 & 0000 & 0111 & 1111 & 0000 & 1111 & 1111 \\
    051 & 0000 & 0000 & 0000 & 1000 & 0000 & 1111 & 0100 \\
    052 & 0000 & 0000 & 0000 & 0000 & 0000 & 0111 & 1011 \\
    053 & 0000 & 0000 & 0000 & 0000 & 0000 & 0000 & 0000 \\
    054 & 0000 & 0000 & 0000 & 0000 & 0000 & 0001 & 0000 \\
    \bottomrule
  \end{tabular}
\end{center}
\clearpage

\begin{center}
  \captionof{table}{Per-task outcome grids, tasks 055--102. Columns follow the blocks of Table~\ref{tab:threeway}: the four Mistral Large runs, then the three Claude Sonnet 5 runs; \emph{S store} and \emph{M store} are the frozen-store consumer cells.}
  \label{tab:grids_b}
  \scriptsize\ttfamily
  \begin{tabular}{lccccccc}
    \toprule
    & \multicolumn{4}{c}{\rmfamily Mistral Large} & \multicolumn{3}{c}{\rmfamily Claude Sonnet 5} \\
    \cmidrule(lr){2-5} \cmidrule(lr){6-8}
    \rmfamily Task & \rmfamily Base & \rmfamily Exp & \rmfamily Instr & \rmfamily S store & \rmfamily Base & \rmfamily Instr & \rmfamily M store \\
    \midrule
    055 & 0000 & 0000 & 0000 & 0000 & 0000 & 0000 & 0000 \\
    056 & 0000 & 0000 & 0000 & 1000 & 0000 & 0001 & 1000 \\
    057 & 0000 & 0000 & 0000 & 0000 & 0111 & 0000 & 1111 \\
    058 & 0000 & 0010 & 1111 & 0100 & 0010 & 0011 & 0000 \\
    059 & 0000 & 0000 & 0000 & 0000 & 0000 & 0001 & 0000 \\
    060 & 0000 & 0000 & 0000 & 1101 & 0000 & 0000 & 0000 \\
    061 & 0000 & 0000 & 0000 & 0000 & 0000 & 0000 & 0010 \\
    062 & 0000 & 0000 & 0000 & 1110 & 0000 & 0000 & 0000 \\
    063 & 0000 & 0000 & 0000 & 0000 & 0000 & 0000 & 0000 \\
    064 & 0000 & 0000 & 0001 & 0000 & 0010 & 0000 & 0000 \\
    065 & 0000 & 0000 & 0000 & 0001 & 0000 & 0010 & 0000 \\
    066 & 0000 & 0000 & 0100 & 1001 & 0000 & 0000 & 0000 \\
    067 & 0000 & 0000 & 0000 & 0001 & 0000 & 0000 & 0000 \\
    068 & 0000 & 0000 & 0000 & 0000 & 0000 & 0000 & 0000 \\
    069 & 0000 & 0000 & 0010 & 0000 & 0000 & 0000 & 0000 \\
    070 & 0000 & 0000 & 0000 & 0000 & 0010 & 0000 & 1110 \\
    071 & 0000 & 0000 & 0000 & 0001 & 0000 & 0000 & 0000 \\
    072 & 0000 & 0000 & 0000 & 0001 & 1000 & 0010 & 0010 \\
    073 & 0000 & 0000 & 0010 & 0001 & 1000 & 0011 & 0010 \\
    074 & 0000 & 0000 & 0000 & 0010 & 0000 & 0000 & 0001 \\
    075 & 0000 & 0000 & 0000 & 1101 & 1000 & 0111 & 0111 \\
    076 & 1100 & 0111 & 0000 & 1110 & 1111 & 1111 & 1111 \\
    077 & 0000 & 0000 & 0000 & 0000 & 0000 & 0000 & 0000 \\
    078 & 0000 & 0000 & 0000 & 0000 & 0000 & 0011 & 0000 \\
    079 & 0000 & 0000 & 0000 & 0000 & 0000 & 0111 & 0000 \\
    080 & 0000 & 0000 & 0000 & 0000 & 0000 & 0000 & 0000 \\
    081 & 0000 & 0000 & 0000 & 0000 & 0000 & 0000 & 0000 \\
    082 & 0000 & 0000 & 0000 & 0000 & 0000 & 0000 & 0000 \\
    083 & 0000 & 0000 & 0000 & 0000 & 0000 & 0000 & 0000 \\
    084 & 0000 & 0000 & 0000 & 0000 & 0000 & 0000 & 0000 \\
    085 & 0000 & 0000 & 0000 & 0000 & 0000 & 0000 & 0000 \\
    086 & 0000 & 0000 & 0000 & 0000 & 0000 & 0000 & 0000 \\
    087 & 0000 & 0000 & 0001 & 0000 & 0000 & 0001 & 0001 \\
    088 & 0000 & 0000 & 0101 & 0000 & 0000 & 0000 & 0011 \\
    089 & 0000 & 0000 & 0100 & 1100 & 1101 & 1101 & 0000 \\
    090 & 0000 & 0000 & 0000 & 0000 & 0000 & 0000 & 0000 \\
    091 & 0000 & 0000 & 0000 & 0000 & 0000 & 0001 & 0000 \\
    092 & 0000 & 0000 & 0000 & 0000 & 0000 & 0011 & 0000 \\
    093 & 0000 & 0000 & 0100 & 1100 & 1001 & 0111 & 1101 \\
    094 & 0000 & 0000 & 0000 & 0101 & 1111 & 1101 & 1111 \\
    095 & 0000 & 0000 & 0000 & 0010 & 0100 & 0111 & 1111 \\
    096 & 0000 & 0000 & 0000 & 0011 & 0000 & 0111 & 0000 \\
    097 & 0000 & 0000 & 0000 & 1101 & 0000 & 0111 & 1100 \\
    098 & 0000 & 0000 & 0001 & 0001 & 0010 & 0000 & 0000 \\
    099 & 0000 & 0000 & 0000 & 1111 & 0000 & 0001 & 1001 \\
    100 & 0000 & 0000 & 0000 & 0000 & 0100 & 0000 & 0000 \\
    101 & 0000 & 0000 & 0000 & 0000 & 0000 & 0001 & 0000 \\
    102 & 0000 & 0000 & 0000 & 0000 & 1-00 & 0000 & 0000 \\
    \bottomrule
  \end{tabular}
\end{center}
\clearpage

\section{Per-condition estimates with intervals}
\label{full_tables_appendix}

Tables~\ref{tab:threeway_full} and~\ref{tab:sonnet_full} reproduce every condition of Table~\ref{tab:threeway} with its 95\% cluster-bootstrap interval over tasks (Appendix~\ref{stats_appendix}), including the hold rates of the independent-trial conditions, which the main table omits: the two no-memory baselines and the two frozen-store transfer consumers. For those conditions the hold statistic measures \textbf{re-draw persistence} rather than retention. With no store, or a store that never changes, nothing carries state between trials, so each trial is an independent draw and the probability that a task passed at trial $t$ passes again at $t+1$ is simply that task's per-trial pass rate. Pooling over tasks conditions on a pass, which weights the estimate towards the tasks the agent passes most often; under independence the pooled statistic estimates $\sum_i p_i^2 / \sum_i p_i$ over the per-task pass rates $p_i$. A value well above the overall success rate therefore signals that passes concentrate on a few consistently solvable tasks. It is the reference level against which the memory conditions' hold rates are read: a learning agent must hold its solutions at a rate above what task-difficulty concentration alone produces.

\begin{center}
  \captionof{table}{The four Mistral Large conditions with intervals. Columns as in Table~\ref{tab:threeway}.}
  \label{tab:threeway_full}
  \small
  \begin{tabular}{lcccccc}
    \toprule
    Condition & pass\textasciicircum{}1 & pass\textasciicircum{}2 & pass\textasciicircum{}3 & pass\textasciicircum{}4 & Solved & Hold rate \\
    \midrule
    No-memory baseline & 0.064 & 0.038 & 0.034 & 0.031 & 13/97 & 0.61 \\
     & \scriptsize$[0.031, 0.108]$ & \scriptsize$[0.009, 0.077]$ & \scriptsize$[0.003, 0.072]$ & \scriptsize$[0.000, 0.072]$ & & \scriptsize$[0.250, 0.833]$ \\
    \addlinespace
    Experience & 0.103 & 0.067 & 0.046 & 0.031 & 16/97 & 0.88 \\
     & \scriptsize$[0.057, 0.157]$ & \scriptsize$[0.029, 0.112]$ & \scriptsize$[0.015, 0.088]$ & \scriptsize$[0.000, 0.072]$ & & \scriptsize$[0.750, 1.000]$ \\
    \addlinespace
    Instruction & \textbf{0.170} & \textbf{0.091} & \textbf{0.057} & 0.031 & \textbf{32/97} & 0.65 \\
     & \scriptsize$[0.116, 0.229]$ & \scriptsize$[0.050, 0.137]$ & \scriptsize$[0.026, 0.098]$ & \scriptsize$[0.000, 0.072]$ & & \scriptsize$[0.471, 0.800]$ \\
    \addlinespace
    Reads Sonnet-built store & 0.289 & 0.172 & 0.124 & 0.093 & 51/97 & 0.65 \\
     & \scriptsize$[0.222, 0.358]$ & \scriptsize$[0.113, 0.237]$ & \scriptsize$[0.070, 0.186]$ & \scriptsize$[0.041, 0.155]$ & & \scriptsize$[0.535, 0.746]$ \\
    \bottomrule
  \end{tabular}
\end{center}

\begin{center}
  \captionof{table}{The three Claude Sonnet 5 conditions with intervals. Columns as in Table~\ref{tab:threeway}. One baseline simulation was lost to an infrastructure error; that task contributes three trials.}
  \label{tab:sonnet_full}
  \small
  \begin{tabular}{lcccccc}
    \toprule
    Condition & pass\textasciicircum{}1 & pass\textasciicircum{}2 & pass\textasciicircum{}3 & pass\textasciicircum{}4 & Solved & Hold rate \\
    \midrule
    No-memory baseline & 0.248 & 0.168 & 0.137 & 0.115 & 40/97 & 0.70 \\
     & \scriptsize$[0.179, 0.323]$ & \scriptsize$[0.105, 0.239]$ & \scriptsize$[0.077, 0.204]$ & \scriptsize$[0.053, 0.181]$ & & \scriptsize$[0.569, 0.810]$ \\
    \addlinespace
    Instruction & \textbf{0.397} & \textbf{0.268} & \textbf{0.206} & \textbf{0.165} & \textbf{62/97} & 0.83 \\
     & \scriptsize$[0.320, 0.474]$ & \scriptsize$[0.196, 0.345]$ & \scriptsize$[0.137, 0.284]$ & \scriptsize$[0.093, 0.247]$ & & \scriptsize$[0.745, 0.904]$ \\
    \addlinespace
    Reads Mistral-built store & 0.314 & 0.230 & 0.191 & 0.165 & 46/97 & 0.78 \\
     & \scriptsize$[0.240, 0.394]$ & \scriptsize$[0.158, 0.309]$ & \scriptsize$[0.121, 0.268]$ & \scriptsize$[0.093, 0.247]$ & & \scriptsize$[0.671, 0.861]$ \\
    \bottomrule
  \end{tabular}
\end{center}

\section{Statistical methodology}
\label{stats_appendix}

\paragraph{Estimators.} pass\textasciicircum{}k uses the benchmark's unbiased estimator: a task with $c$ successes in $n=4$ trials contributes $\binom{c}{k}/\binom{n}{k}$, and the statistic is the mean over tasks. For $k=1$ this equals the pooled success rate; for $k=n$ it is the indicator of solving every trial. Per-trial success rates are pooled over tasks at a fixed trial index. The hold rate pools all transitions $(t, t{+}1)$ in which trial $t$ passed and reports the fraction in which trial $t{+}1$ also passed. The floor-conversion statistic uses stratum labels fixed once from the baseline arm's actual results, per the rule in Section~\ref{protocol_section}; the labels are never re-derived during resampling.

\paragraph{Interval estimation.} Trials within a task share the task's difficulty and are correlated, so the task is the sampling unit. Intervals are cluster-bootstrap percentile intervals: $B = 10{,}000$ replicates, each resampling the 97 task indices with replacement and carrying every sampled task's complete trial vectors. The same resampled index set is applied to every condition within a replicate, preserving the pairing of conditions on tasks; contrasts between conditions are computed inside each replicate and their percentiles reported. Replicates in which a statistic is undefined (a hold rate with no qualifying transitions) are excluded from that statistic's percentiles and counted; the count is reported when it is non-zero. The generator seed is fixed and published with the analysis code, which reproduces every reported number, point estimates and intervals, from the released run data.

\end{document}